\documentclass[10pt,twocolumn,letterpaper]{article}

\usepackage{wacv}
\makeatletter
\@namedef{ver@everyshi.sty}{}
\makeatother
\usepackage{pgfplots}

\usepackage{times}
\usepackage{epsfig}
\usepackage{graphicx}
\usepackage{amsmath}
\usepackage{amssymb}

\usepackage{caption}
\usepackage{adjustbox}
\usepackage{wrapfig}
\usepackage{xcolor}
\usepackage{multirow}
\usepackage{booktabs}
\usepackage{url}
\usepackage{subfigure}
\usepackage{enumitem}

\def\wacvPaperID{1196} 

\wacvfinalcopy 

\ifwacvfinal
\def\assignedStartPage{1} 
\fi

\ifwacvfinal
\usepackage[breaklinks=true,bookmarks=false]{hyperref}
\else
\usepackage[pagebackref=true,breaklinks=true,colorlinks,bookmarks=false]{hyperref}
\fi

\ifwacvfinal
\else
\fi

\begin{document}

\title{Dissected 3D CNNs: Temporal Skip Connections for \\
       Efficient Online Video Processing}

\author{\parbox{16cm}{\centering
		{\large Okan K\"op\"ukl\"u$^1$, \hspace{0.2cm} Stefan H\"ormann$^1$, \hspace{0.2cm} Fabian Herzog$^1$, \hspace{0.2cm} Hakan Cevikalp$^2$, \hspace{0.2cm} Gerhard Rigoll$^1$}\\
		{\normalsize
			\vspace{0.2cm}
			$^1$ Technical University of Munich \\
			$^2$ Eskisehir Osmangazi University}}
}

\maketitle

\begin{abstract}

Convolutional Neural Networks with 3D kernels (3D-CNNs) currently achieve state-of-the-art results in video recognition tasks due to their supremacy in extracting spatiotemporal features within video frames. There have been many successful 3D-CNN architectures surpassing the state-of-the-art results successively. However, nearly all of them are designed to operate offline creating several serious handicaps during online operation. Firstly, conventional 3D-CNNs are not dynamic since their output features represent the complete input clip instead of the most recent frame in the clip. Secondly, they are not temporal resolution-preserving due to their inherent temporal downsampling. Lastly, 3D-CNNs are constrained to be used with fixed temporal input size limiting their flexibility. In order to address these drawbacks, we propose dissected 3D-CNNs, where the intermediate volumes of the network are dissected and propagated over depth (time) dimension for future calculations, substantially reducing the number of computations at online operation. For action classification, the dissected version of ResNet models performs 77-90\% fewer computations at online operation while achieving $\sim$5\% better classification accuracy on the Kinetics-600 dataset than conventional 3D-ResNet models. Moreover, the advantages of dissected 3D-CNNs are demonstrated by deploying our approach onto several vision tasks, which consistently improved the performance.


\end{abstract}
\section{Introduction}
\label{sec:intro}

Convolutional Neural Networks (CNNs) have dominated the majority of computer vision tasks ever since AlexNet \cite{krizhevsky2012imagenet} won the ImageNet Challenge (ILSVRC 2012 \cite{russakovsky2015imagenet}). In order to harness a similar performance as 2-dimensional (2D) CNNs achieved on image-based tasks, 3-dimensional (3D) CNNs have been proposed by adding an additional depth dimension to convolutional and pooling layers. However, 3D-CNNs require significantly more parameters and computations at inference time compared to their 2D counterparts making them more challenging to train and prone to overfitting. The overfitting problem is resolved with the availability of large scale video datasets, such as Kinetics \cite{carreira2017quo}, Sports-1M \cite{karpathy2014large}. Nevertheless, computational cost remains as the biggest drawback of 3D-CNNs. 


\begin{figure}[t!]
\centering
\vspace{0.2cm}
\includegraphics[width=1.0\linewidth]{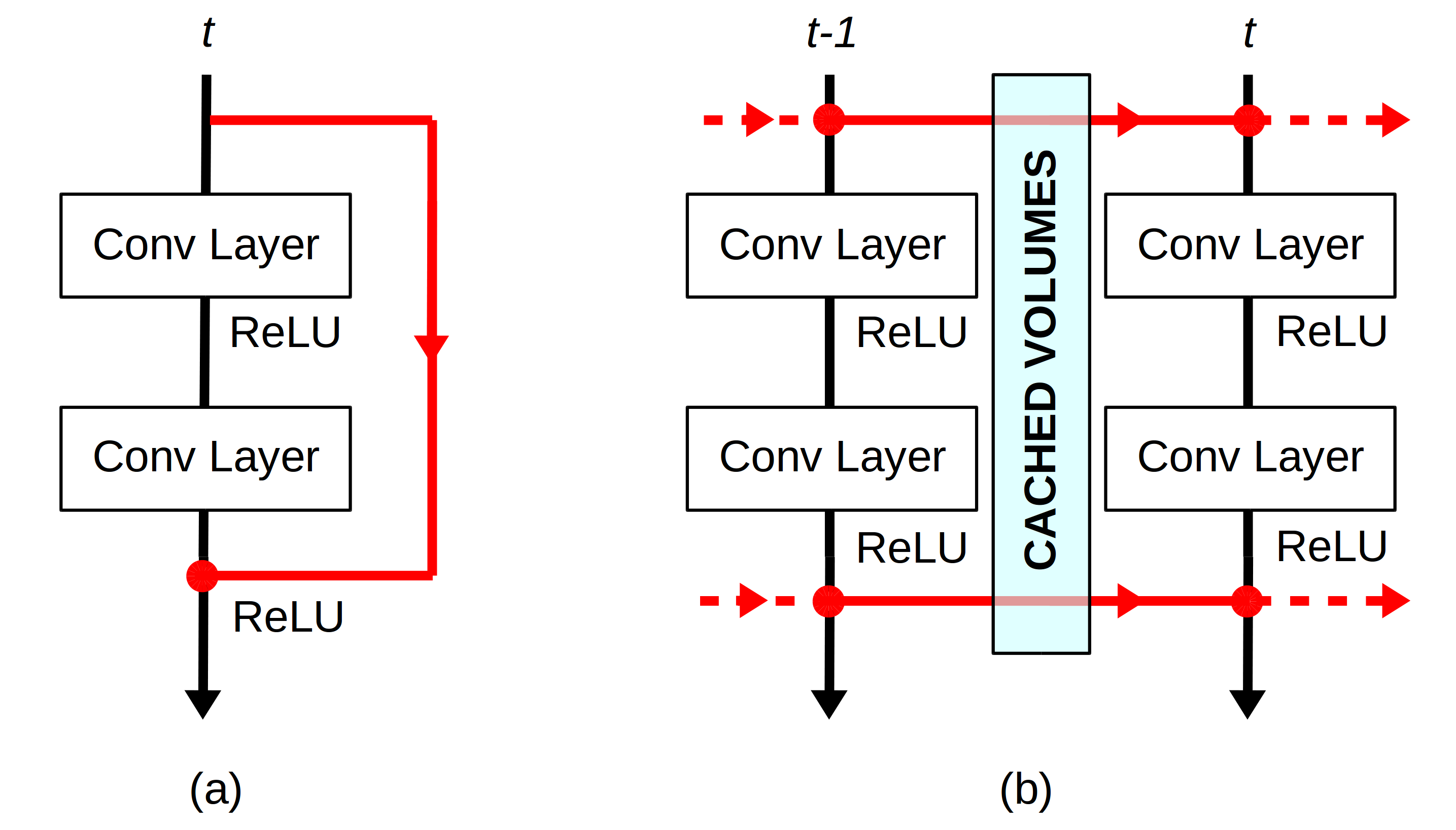}
\caption{Comparison of spatial skip connections (a) first proposed in \cite{he2016deep} and temporal skip connections (b) proposed in this work. At every iteration, only the computations for the most recent frame are performed. Afterwards, intermediate volumes from the skip connections are cached to be used for the next iteration. This way, recomputation of previous frames is saved. Skip connections are denoted with red lines.}
\vspace{0.2cm}
\label{fig:spatial_vs_temporal}
\end{figure}

Currently, the primary trend in video recognition tasks is to increase network performance by building deeper and wider 3D-CNN architectures \cite{hara2018can, feichtenhofer2018slowfast, carreira2017quo}. However, these architectures are typically designed to operate offline, ignoring the requirements of online operation. Firstly, most of the 3D-CNNs deploy temporal downsampling to reduce the computational cost at the later stages of the network and provide translation invariance (in the time dimension) to the internal representation. This causes the network to become non-dynamic, which is of utmost importance for online operation. Moreover, the resulting network is not temporal resolution-preserving. Secondly, 3D-CNNs are typically built to work with a fixed number of input frames. Therefore, online operating frameworks usually use 3D-CNNs in a sliding window, either with small temporal stride \cite{kopuklu2019real, kopuklu2019talking} or larger stride \cite{molchanov2016online}. In the former case, there is a severe resource waste due to reprocessing frames in the overlapping regions, which are already processed in the previous timestamps. In the latter case, there is an information loss since relations between some of the frames are not exploited. These issues make most of the 3D-CNNs unsuitable for online operation.

\begin{figure}[t!]
\centering
\vspace{0.2cm}
\includegraphics[width=1.02\linewidth]{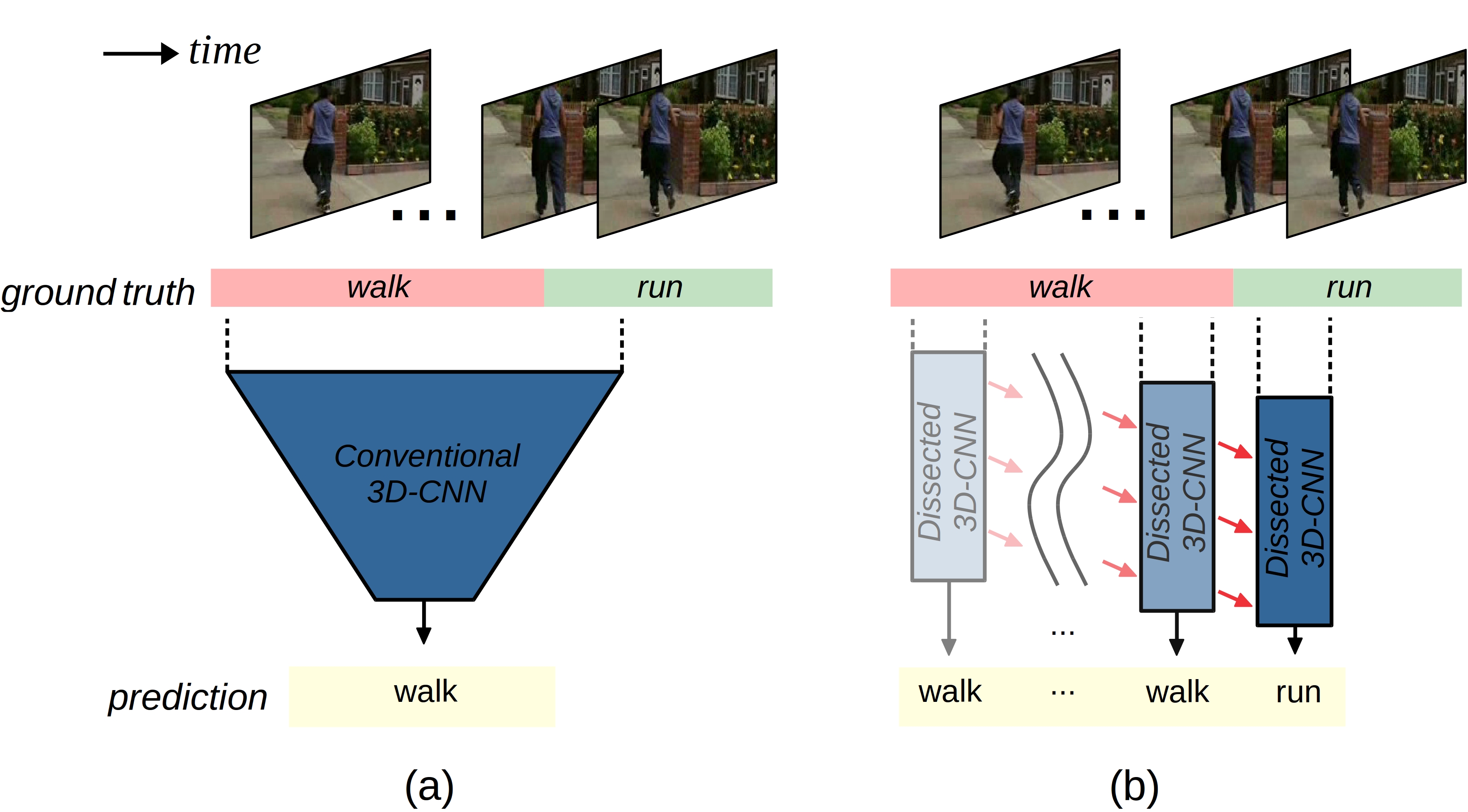}
\caption{Comparison of Conventional 3D-CNNs (a) and Dissected 3D-CNNs (b) at online operation. Conventional 3D-CNNs work with fixed number of input frames and their predictions can be triggered by any frame in the input clip. Therefore, conventional 3D-CNNs are \textit{non-dynamic} and although a new action starts in the video stream, they can continue predicting previous action, as illustrated in (a). On the other hand, Dissected 3D-CNNs processes video with each new coming frame and update their prediction dynamically, as illustrated in (b). Red lines in (b) denotes temporal skip connections.}
\label{fig:dynamic_operation}
\end{figure}

In order to address the limitations mentioned above, we propose a novel 3D-CNN architecture, Dissected 3D-CNNs (D3D), by incorporating temporal skip connections. Skip connections are first proposed in ResNets \cite{he2016deep} to overcome the issue of vanishing/exploding gradients and to enhance gradient propagation for deep architectures. Spatial skip connections, which are depicted in Fig.~\ref{fig:spatial_vs_temporal}~(a), can be in the form of summation \cite{he2016deep} or concatenation \cite{huang2017densely, ma2018shufflenet}. As opposed to spatial skip connections, we propose temporal skip connections to create a network for efficient online operation. The general idea of the proposed architecture is depicted in Fig.~\ref{fig:spatial_vs_temporal}~(b). Intermediate volumes are always stored in a cache, and only the computations for the new available frame are performed at each iteration. After the computations, the previously cached volumes are replaced with the most recent intermediate feature volumes coming from the skip connections. This way, the volumes in Dissected 3D-CNN architecture are propagated without calculating them repeatedly. We incorporate 3D convolutions since we apply concatenation operation in the depth dimension at the skip connections. Although summation is also possible at temporal skip connections, we will show in our ablation study that temporal information is lost with the summation operation, which leads to inferior results. Moreover, spatial skip connections are still applicable on top of temporal skip connections.

The main motivation of this work is to provide a 3D-CNN architecture which satisfies the requirements of online operation. Out of many, two most important requirements are (i) dynamic operation and (ii) reduced computational complexity. We refer the term \textit{dynamic} as deployed architecture's ability to adapt its output according to the new coming video frames. Fig. \ref{fig:dynamic_operation} illustrates the comparison of conventional 3D-CNNs and Dissected 3D-CNNs at online operation. Conventional 3D-CNNs are non-dynamic since the final decision of the network might be triggered by any previous frame in the input clip, not due to the latest introduced frame. This is specifically critical for the online recognition of actions which is performed in very short time intervals, such as Driver Micro Hand Gestures (DriverMHG) dataset \cite{kopuklu2020driver}. On the other hand, Dissected 3D-CNNs need to process only the most recent frame at online operation since they can leverage the previously computed intermediate volumes via a caching mechanism. Consequently, Dissected 3D-CNNs can update their predictions according to the new coming frame and hence operated dynamically while enjoying reduced computational complexity.

To obtain the network’s final decision, dissected 3D-CNN architecture still needs a spatiotemporal modeling mechanism at the end. Although the conventional way of using a fully connected layer is a valid option, a Recurrent Neural Network (RNN) block can also be applied. The RNN block makes the D3D architecture independent of the number of input frames and performs better, as shown in our ablation study. Overall, Dissected 3D-CNNs bring the following advantages:
\begin{enumerate}
    \item D3Ds provide frame-level features.
    \item D3Ds operate at any number of input frames.
    \item Any 3D-CNN architecture can be converted to its dissected version\footnote{D3D is a general term referring all the dissected 3D-CNN architectures which employ temporal skip connections. For the dissected version of a specific network architecture, we use the prefix `D' (e.g., the dissected version of ResNet-18 is denoted as D-ResNet-18).}.
    \item A large number of computations are saved at online operation. Dissected versions of ResNet-18,50,101 perform 77-90\% less computation at online operation while achieving $\sim$5\% better classification accuracy compared to conventional ResNet models on \mbox{Kinetics-600} dataset. 
    \item Any frame-level task can leverage from D3D architecture if the frames are obtained from continuous video streams.
\end{enumerate}

The remaining part of the paper is organized as follows: In Section~\ref{sec:relwork}, the related work is presented. Section~\ref{sec:method} introduces the D3D architecture and elaborates training and evaluation processes. Section~\ref{sec:exp} presents experiments and results on various video-based computer vision tasks. Finally, Section~\ref{sec:con} concludes the paper.


\section{Related Work}
\label{sec:relwork}

\textbf{Video-based computer vision tasks.} The proposed D3D is designed to propagate spatiotemporal information with frame-level correspondence. Therefore, any task requiring to process continuous video streams can benefit from D3D including action/activity recognition with datasets UCF-101 \cite{soomro2012ucf101}, HMDB \cite{Kuhne:ICCV:2011}, Kinetics \cite{carreira2017quo}; video object detection task such as ImageNet VID \cite{russakovsky2015imagenet}; spatiotemporal action localization task such as Atomic Visual Actions (AVA) dataset \cite{gu2018ava}; video object tracking (VOT) task such as \cite{VOT_TPAMI, WuLimYang13}; multi-object tracking (MOT) such as \cite{dendorfer2019cvpr19}; video person re-identification task such as MARS (Motion Analysis and Re-identification Set) dataset \cite{zheng2016mars}; gait recognition task such as Casia-B dataset \cite{yu2006framework}; video face recognition such as YouTube Faces \cite{wolf2011YTF} and many other tasks. Currently, state-of-the-art architectures either  use only spatial content by processing the input frame-by-frame ignoring the temporal content \cite{cevikalp2019visual, wang2019fast, bergmann2019tracking} or utilize offline-trained 3D-CNN architectures in a non-dynamic way \cite{kopuklu2019real, athar2020stem}. By utilizing our proposed D3D architecture, all video-based computer vision tasks can incorporate temporal information.

\vspace{0.4cm}
\noindent \textbf{3D-CNN architectures.} Ji \etal \phantom{a}propose a 3D-CNN architectures for the first time in \cite{ji20123d}. Ever since then, there have been plenty of 3D-CNN architectures to achieve better accuracies at video classification tasks such as C3D \cite{tran2015learning}, I3D \cite{carreira2017quo}, R(2+1)D \cite{tran2018closer}, P3D \cite{qiu2017learning}, SlowFast \cite{feichtenhofer2018slowfast}, etc. The effect of dataset size is investigated in \cite{hara2018can} together with the performance of widely-used architectures such as ResNet \cite{he2016deep}, DenseNet \cite{huang2017densely}, ResNext \cite{xie2017aggregated}. In \cite{kopuklu2019resource}, 3D versions of popular resource-efficient architectures are investigated for video classification tasks. X3D is proposed in \cite{feichtenhofer2020x3d}, which is a spatiotemporal architecture expanded from a tiny spatial network by multiple axes in space, time, width and depth in order to ensure good computation/accuracy trade-off. However, all these architectures are designed for offline operation and do not meet the requirements for online operation, as they operate with a fixed number of input frames. Moreover, the number of floating point operations (FLOPs) is in the order of 10s-100s GFLOPs at inference time, which is too costly for online operation.

\begin{table*}[t!]
\begin{minipage}[b]{0.607\linewidth}
\centering
\includegraphics[height=6.5cm]{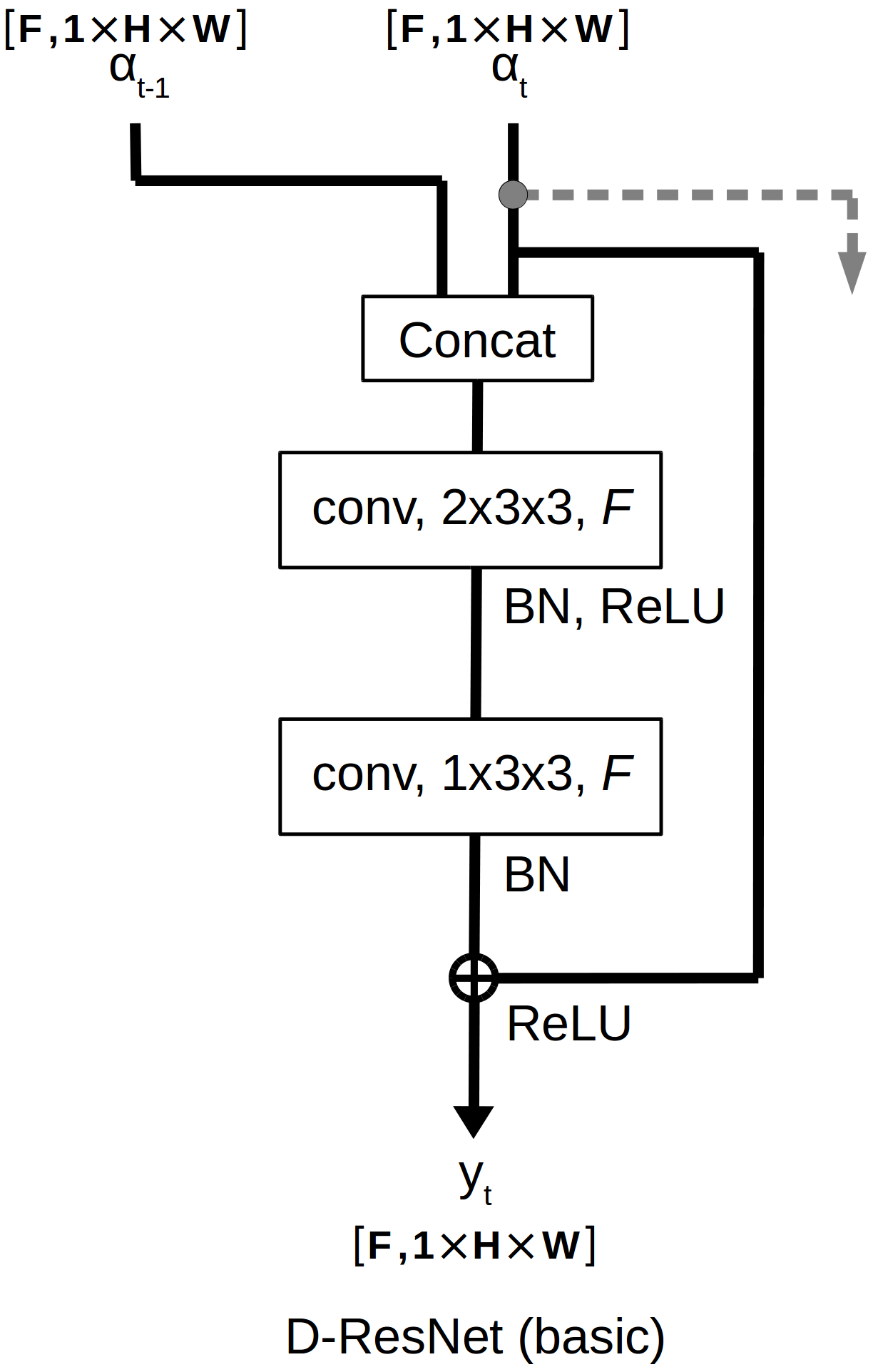} \hspace{0.4cm} \includegraphics[height=6.5cm]{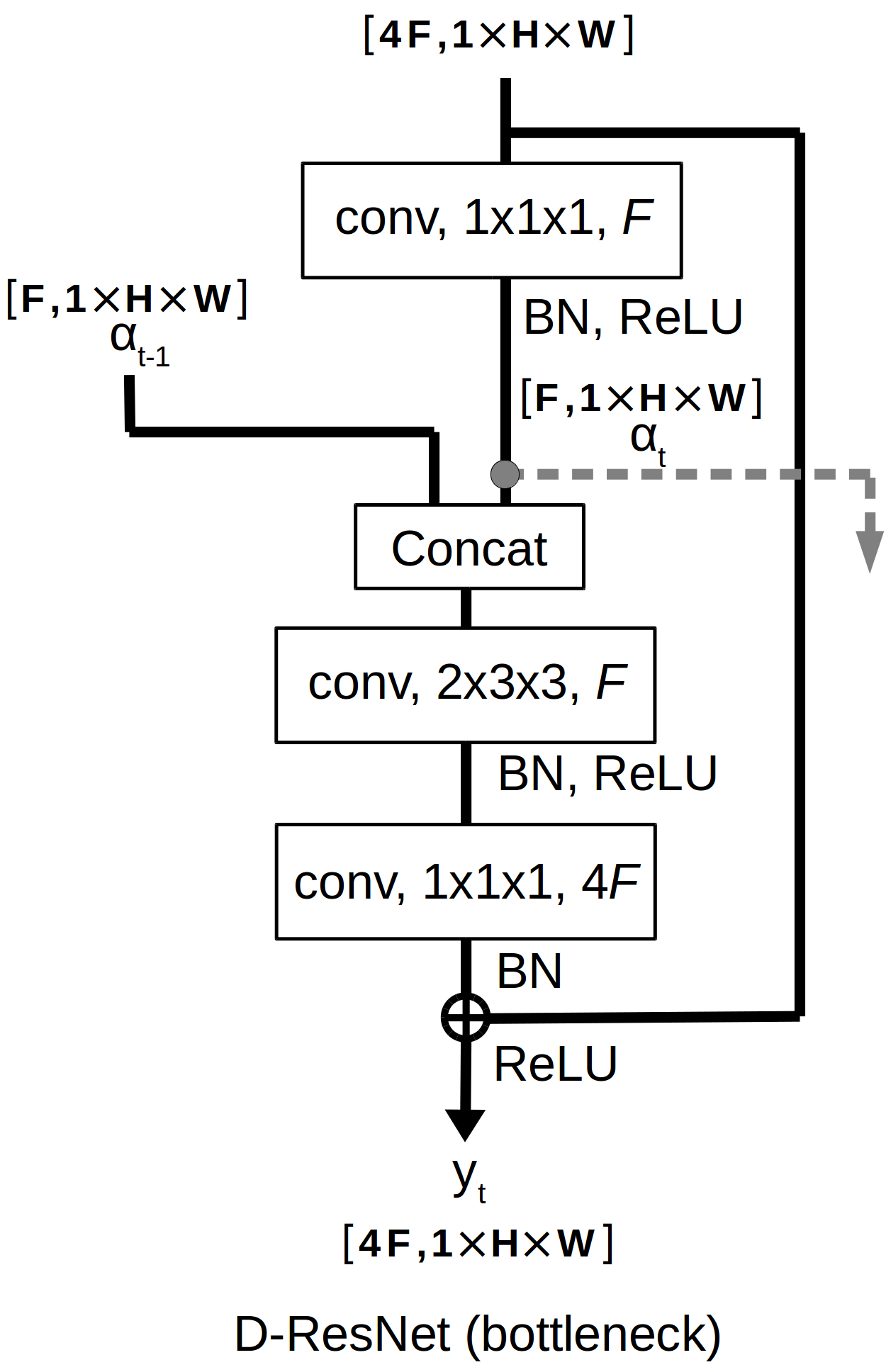}
\captionof{figure}{Basic and bottleneck blocks used in ResNet architecture. $F$, $BN$, and $ReLU$ denote the number of feature maps (i.e. channels), batch normalization \cite{ioffe2015batch}, and rectified linear unit, respectively. $Concat$ denotes concatenation at depth dimension while $\oplus$ denotes to element-wise addition.}
\label{fig:blocks}
\end{minipage}
\hfill
\begin{minipage}[b]{0.375\linewidth}
\centering
\begin{adjustbox}{width=0.97\textwidth}
\vspace{-4cm}
	\begin{tabular}{l|c|c}
		\hline \rule{0pt}{12pt}
		\textbf{Layer}    & \textbf{ResNet-18}  & \textbf{ResNet-\{50,101\}} \\[0.1cm]  \hline \rule{0pt}{12pt} 
		block      & basic   & bottleneck   \\[0.1cm] \hline \rule{0pt}{12pt} 
		conv1      & \multicolumn{2}{c}{conv(3$\times$7$\times$7), stride (1, 2, 2)} F:64  \\[0.1cm] \hline \rule{0pt}{12pt}
		pool       & \multicolumn{2}{c}{ MaxPool(1$\times$3$\times$3), stride (1, 2, 2)}  \\[0.1cm] \hline \rule{0pt}{12pt}
		conv2\_x   & N:2, F:64   & N:3, F:64   \\[0.1cm] \hline \rule{0pt}{12pt} 
		conv3\_x   & N:2, F:128 & N:4, F:128    \\[0.1cm] \hline \rule{0pt}{12pt}
		conv4\_x   & N:2, F:256 & N:\{6, 23\}, F:256    \\[0.1cm] \hline \rule{0pt}{12pt}
		conv5\_x   & N:2, F:512 & N:3, F:512   \\[0.1cm] \hline \rule{0pt}{12pt}
		conv\_last & ---   & \begin{tabular}[c]{@{}c@{}}conv(1$\times$1$\times$1), \\ stride (1, 1, 1), F:512 \end{tabular}  \\[0.1cm] \hline \rule{0pt}{12pt}
		         & \multicolumn{2}{c}{\begin{tabular}[c]{@{}c@{}}global average pooling,\\ spatiotemporal modeling\end{tabular}} \\[0.2cm] \hline
	\end{tabular}
\end{adjustbox}
\caption{Dissected ResNet architectures. $F$ is the number of feature channels corresponding in Fig. \ref{fig:blocks}, and $N$ refers to the number of blocks in each layer.}
\label{table:arch_details}
\end{minipage}
\end{table*}

\vspace{0.4cm}
\noindent  \textbf{Online video processing architectures.} For video activity recognition and description, a Long-term Recurrent Convolutional Networks (LRCN) are proposed in \cite{donahue2015long}, which extracts features from video frames by a 2D-CNN and applies LSTM as spatiotemporal modeling mechanism. However, frame-wise features are extracted independent from each other, hence pixel-wise motion information cannot be extracted with LRCN. A temporal shift module (TSM) is introduced in \cite{lin2019tsm}, which can be inserted into 2D CNNs to shift part of the channels along the temporal dimension in order to facilitate information exchange among neighboring frames. TSM is a dynamic architecture and causal version of it can be created by applying only single-sided shifting operation. However, TSM again lacks capturing pixel-wise motion information since it operates with 2D CNNs. For gesture recognition, Molchanov \etal \phantom{a}propose to use 3D-CNN to extract features followed by an LSTM for online recognition \cite{molchanov2016online}. However, this approach is \textit{non-dynamic} since the 3D-CNN processes non-overlapping 8-frame clips. K\"op\"ukl\"u \etal \phantom{a}propose to use a two-level hierarchical framework for online gesture recognition \cite{kopuklu2019real}. This architecture is again \textit{non-dynamic} since the detector also takes 8-frame clips with a sliding window. For spatiotemporal action localization task, \cite{singh2017online, kalogeiton2017action} propose to use a detector to obtain frame-level detections and create action tubes with further post-processing. However, these methods make use of optical flow modality in order to incorporate motion information, which requires a substantial amount of computation.   In \cite{kopuklu2019you}, the YOWO (You Only Watch Once) architecture is proposed, where spatiotemporal and fine-spatial features are concurrently extracted via a 3D-CNN and 2D-CNN, and actions are detected on the key-frame. YOWO is a \textit{dynamic} architecture in this regard. However, YOWO is not watching once since it operates using a sliding window for continuous videos, and 15 frames of a 16-frame clip have already been processed (\textit{watched}) in the previous step. So there is a serious amount of repetitive  computation at online operation, which can be avoided. The closest work to ours is \cite{singh2019recurrent}, in which Singh \etal \phantom{a}propose to decompose a 3D convolutional block into a 2D spatial convolution followed by a recurrent unit for temporal modeling. However, in this work, convolutions are performed in 2D (i.e. depth dimension of the convolutional kernels are always 1) and temporal information is captured only with recurrent units. Moreover, putting a recurrent unit at each layer of the network is too costly in terms of computation complexity. To the best of our knowledge, D3D is the first architecture proposing to propagate intermediate volumes of the complete 3D-CNN architecture to reduce the computational complexity during online operation. 
\section{Methodology}
\label{sec:method}

In this section, we first elaborate on the D3D architecture details, which reduces the computational complexity substantially during online operation. Secondly, we mention possible options for spatiotemporal modeling. Finally, training details are described.

\subsection{Dissected 3D-CNN Architecture}

In order to demonstrate the advantages of the proposed D3D architecture, we have created the dissected version of the ResNet family (named as D-ResNet) and compared its performance with the conventional 3D-ResNet family as in \cite{kopuklu2019resource}. The details of the proposed D-ResNet models and corresponding basic and bottleneck blocks are shown in Table~\ref{table:arch_details} and Fig.~\ref{fig:blocks}, respectively. Similar to original ResNet architecture \cite{he2016deep}, spatial downsampling is performed at $conv1$, $pool$, $conv3\_1$, $conv4\_1$, and $conv5\_1$ with a stride of 2. No temporal downsampling is employed. Depending on the number of frames used at inference time (only current frame or current frame with two previous frames), convolution kernel for $conv1$ layer is selected as (1$\times$7$\times$7) or (3$\times$7$\times$7). Unlike the 3D-ResNet architectures, we reduced the depth dimension of the initial convolutional layer of the basic block and the middle convolutional layer of the bottleneck block to 2 since we cache only previous intermediate volumes. We also modify the second convolutional layer of the basic block and set its depth dimension to 1. Excluding spatiotemporal modeling mechanisms, these modifications lead to parameter reduction of $\sim$50\%  on D-ResNet-18 and $\sim$23\% on D-ResNet-50,101 compared to conventional 3D-ResNet architectures.

\begin{figure}[t!]
\centering
\includegraphics[height=7.5cm]{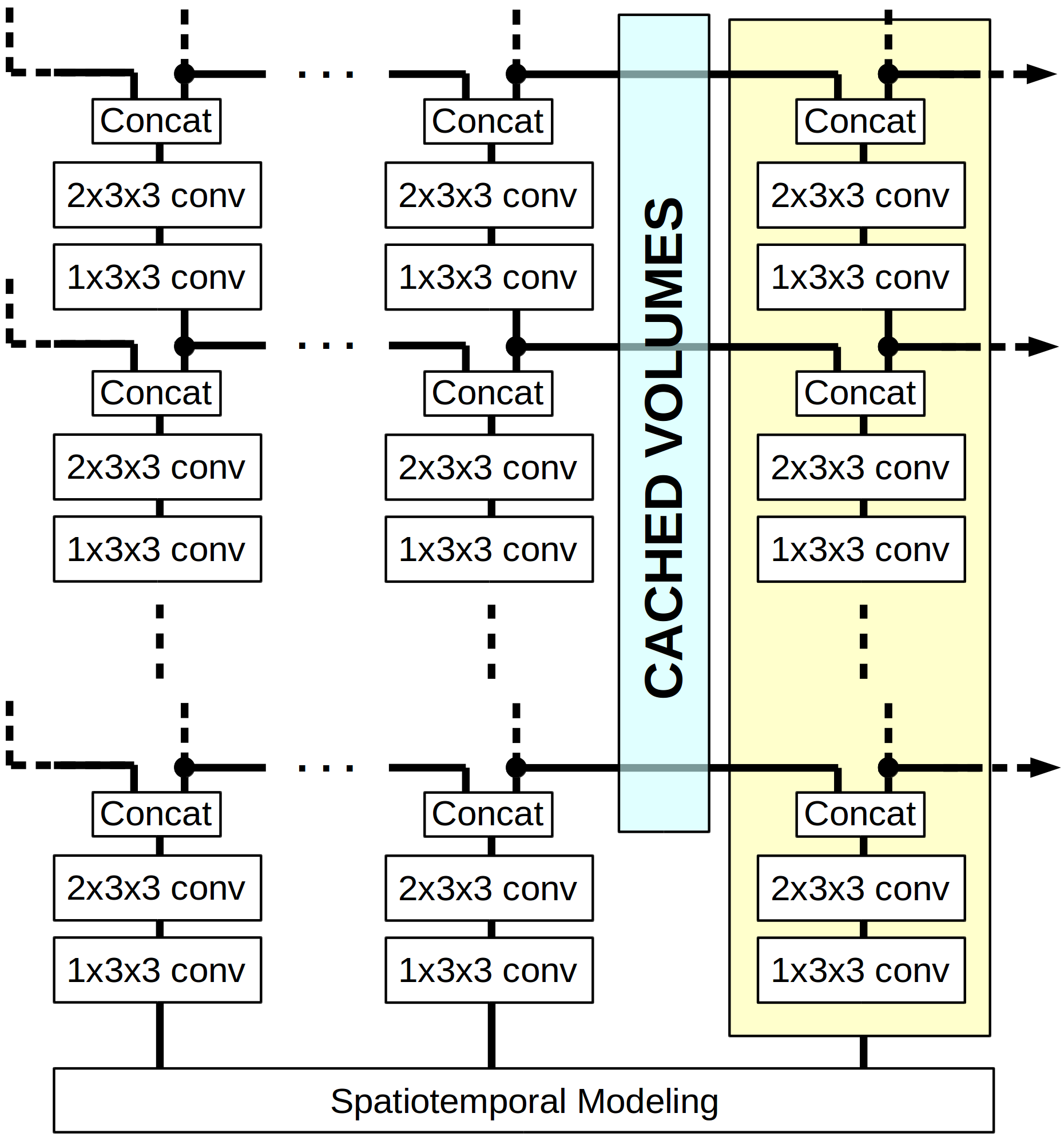}
\caption{Proposed Dissected 3D-CNN architecture using basic D-ResNet block. At the online operation time, intermediate volumes from the previous timestamp is stored in a cache (blue region), and only the computations for the current frame frame are performed (yellow region). Spatial skip connections are excluded for the sake of simplicity.}
\label{fig:D3D_arch}
\end{figure}

An illustration of Dissected 3D-CNN architecture with basic D-ResNet block is shown in Fig. \ref{fig:D3D_arch}. The primary motivation to create such an architecture is to avoid the recomputation of already processed frames of the video stream during online operation. For that, intermediate volumes of the architecture are stored in a cache (blue region in Fig. \ref{fig:D3D_arch}) and used at inference. Throughout the inference, previous intermediate volumes in the cache are replaced with the current ones to be used in the next iteration. Therefore, only the computations within the yellow region in Fig. \ref{fig:D3D_arch} are performed at online operation. Moreover, the designed D3D architecture does not employ (i) temporal downsampling and (ii) padding from right to ensure dynamic online operation. 

At inference time, only the current frame or current frame with previous two frames are passed to the network depending on the task at hand. The reason of leveraging previous two frames is to capture pixel-wise motion information, which is critical for motion intensive datasets such as Jester dataset \cite{materzynska2019jester}. At the first iteration, same padding is applied at $concat$ operations since the cache for the intermediate volumes is empty. For D-ResNet-50,101 architectures, an additional $conv\_last$ block is used in order to reduce the output feature dimension from 2048 to 512. So, all D-ResNet architectures produce a 512-dimensional feature vector for every frame. After obtaining frame-level features, a spatiotemporal modeling mechanism is required to produce class-conditional probabilities, which is explained in the next section.

\subsection{Spatiotemporal Modeling Mechanism}

The typical approach for spatiotemporal modeling is to conclude the network with a fully connected (fc) layer. This approach is also how we trained our architectures from scratch. However, the fc layer at the end of the network requires a fixed number of frames as input. Moreover, the dynamicity condition of the architecture is not met since the decision is made with all output features coming from each frame in the clip.

\begin{figure}[t!]
\centering
\includegraphics[height=7.5cm]{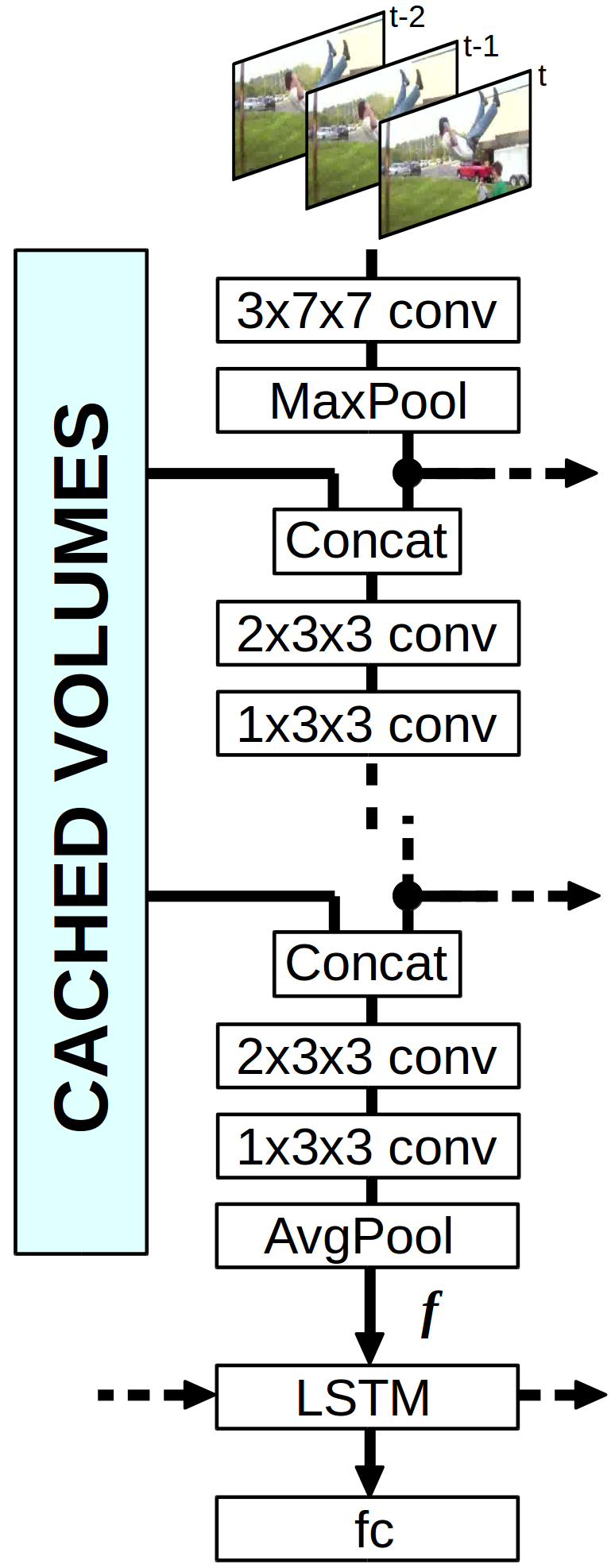}
\caption{\mbox{D-ResNet-18} architecture with LSTM spatiotemporal modeling mechanism. Spatial skip connections are excluded for the sake of simplicity.}
\vspace{-0.1cm}
\label{fig:dynamic_d3d}
\end{figure}

In order to achieve a dynamic architecture, we have considered two popular RNN blocks: Long Short-Term Memory (LSTM) \cite{hochreiter1997long} and gated recurrent unit (GRU) \cite{chung2014empirical}. However, joint end-to-end training of the feature extraction and RNN blocks is not feasible due to the computational and memory complexity of back-propagating through the long video, as described in \cite{wu2019long}. To this end, we have extracted the output features $f$ (before the fully connected layer - see Fig.~\ref{fig:dynamic_d3d}) of all video frames for the training and test set and trained the recurrent blocks separately. For example, each video in the Kinetics dataset lasts around 10 seconds, which makes 250 frames if the video is recorded with 25 fps. After applying the recurrent block, an fc layer is used at the last output of the recurrent block to map the hidden feature map to the number of classes. We have named the resulting network as purely dynamic \mbox{D-ResNet-18} architecture since the network produces a decision using the most recent frame at every iteration. D-ResNet-18 architecture with LSTM spatiotemporal modeling mechanism is shown in Fig. \ref{fig:dynamic_d3d}. In the experiments section, we will validate the advantages of recurrent spatiotemporal modeling techniques.

\subsection{Implementation Details}
\label{sec:imp_details}

\vspace{0.25cm}
\textbf{Learning}: We initially train our D3D architectures with fc layer at the end. 19 frames are fed to the network, but only the last 16 output features are used for loss computation. The reason of feeding 3 frames more is to initialize the cached intermediate volumes properly. Stochastic Gradient Descent (SGD) is applied with standard categorical cross-entropy loss as an optimizer. The largest fitting batch size is selected for mini-batch size, which is typically in the order of 128 clips. The networks are trained from scratch with a learning rate initialized with 0.1 and reduced 3 times with a factor of 10$^{-1}$ when the validation loss converges. For temporal augmentation, clips are selected from a random position in the video. For spatial augmentation, clips are selected from a random spatial position with a randomly selected scale from  \{1, $\frac{1}{2^{1/4}}$, $\frac{1}{2^{3/4}}$, $\frac{1}{2}$\} in order to perform multi-scale cropping as in \cite{hara2018can}. For the case of stacked fc layers, hidden dimension of 1024 is applied.

\begin{table*}[t!]
    \centering
    \begin{tabular}{lccccc}
        \specialrule{.15em}{.1em}{.1em}
        \textbf{Model} & \textbf{Skip Connection} & \textbf{Params} & \textbf{MFLOPs} & \textbf{St-Modeling} & \textbf{Accuracy (\%)}   \\ 
        \specialrule{.15em}{.1em}{.1em}
        D-ResNet-18    & None           & 15.94M   & 546     & fc  & 58.74    \\
        D-ResNet-18    & Summation      & 15.94M   & 546     & fc  & 58.40     \\
        D-ResNet-18    & Concatenation  & 20.66M   & 747     & fc  & 61.41    \\
        \specialrule{.15em}{.1em}{.1em}
    \end{tabular}
    \caption{Performance comparison for different temporal skip connections at online operation on the Kinetics-600 validation set. }
	\label{tab:comparison_sk}
\end{table*}

\begin{table*}[t!]
\begin{minipage}[b]{0.485\linewidth}
\centering
\begin{adjustbox}{width=0.95\textwidth}
\begin{tikzpicture}
      \begin{axis}[
        width=10cm,
        height=4cm,
        xlabel=Training clip length (number of frames),
        ylabel=Accuracy ($\%$),
        grid=both,
        legend columns=2,
        legend style={at={(0.5,0.30)},anchor=north}]
        \addplot coordinates {
            (16,   50.22)
            (32,   50.42)
            (64,   53.35)
            (96,   55.97)
            (128,  57.67)
            (160,  59.27)
            (192,  60.32)
            (224,  61.23)
            (256,  62.02)
        };
      \end{axis}
\end{tikzpicture}
\end{adjustbox}

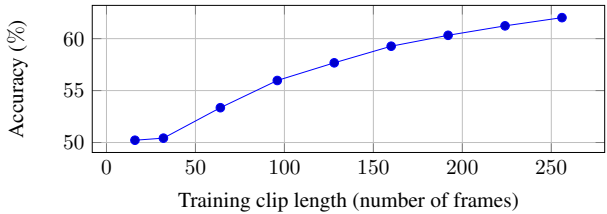
\captionof{figure}{Influence of using different clip lengths at training on the accuracy of the Kinetics-600 validation set when training a D-ResNet-18-lstm.}
\label{fig:clip_lengths}
\end{minipage}
\hfill
\begin{minipage}[b]{0.485\linewidth}
\centering
\begin{adjustbox}{width=0.60\textwidth}
	\begin{tabular}{c|c|c|c}
		\hline \rule{0pt}{12pt}
		\textbf{Layer} & \textbf{1-layer}  & \textbf{2-layer} & \textbf{3-layer}\\[0.1cm]  \hline \rule{0pt}{12pt} 
		GRU            & 61.08             & 61.43            & 61.38   \\[0.1cm] \hline \rule{0pt}{12pt} 
		LSTM           & 61.10             & \textbf{62.02}   & 60.83  \\[0.1cm] \hline  \rule{0pt}{12pt} 
		fc             & 61.41             & 61.25            & 61.22   \\[0.1cm] \hline 
	\end{tabular}
\end{adjustbox}
\vspace{0.4cm}
\caption{Accuracy on the Kinetics-600 validation set for different spatiotemporal modeling mechanisms using D-ResNet-18 architecture.}
\label{table:comp_st-modeling}
\end{minipage}
\end{table*}

For the training of the RNN blocks, we again use SGD with identical learning rates. However, we apply different augmentation schemes. First, the number of input features is selected randomly between [16, '\textit{number of frames in the video}'] and padded with zero to obtain a fixed size of input for all videos. In this way, the RNN blocks can learn all \textit{short}-, \textit{medium}- and \textit{long}-range dependencies. Moreover, videos are temporally down-sampled by 2, 3 and 4 with probabilities of 30\%, 14\% and 11\%, respectively. We also replaced random parts of the input features with noise to enable RNN blocks to ignore unrelated parts of the input. In order to increase regularization, we  also leverage Gaussian noise with zero mean and 0.005 variance at the input features and 0.3 dropout at the hidden layers of RNN blocks. For the hidden layers of RNN blocks, dimension is set to 1024. 

\vspace{0.5cm}
\noindent \textbf{Recognition:} Clips are selected by a sliding window with stride of 1 over the complete video for fc spatiotemporal modeling. Afterwards, class scores are averaged for all the clips. For RNN blocks, the complete input is fed to the network and the last output of the RNN block is used for the final prediction.  

\vspace{0.25cm}
\noindent \textbf{Implementation:} Network architectures are implemented in PyTorch. Our code and pretrained models will be made publicly available\footnote{https://github.com/okankop/Dissected-3D-CNNs}.

\section{Experiments}
\label{sec:exp}

\subsection{Video Activity Recognition On Trimmed Datasets}

We perform detailed ablation study on Kinetics-600 \cite{carreira2018short} dataset. Kinetics-600 contains trimmed YouTube clips with an average duration of 10 seconds belonging to 600 different categories. For our evaluations, we have used the validation set of Kinetics-600 dataset since test set is not publicly available.

\vspace{0.1cm}
\noindent \textbf{\textit{Comparison of different temporal skip connection operations:}} We first compare the performance of different temporal skip connection operations. Table \ref{tab:comparison_sk} shows the comparison of applying summation, concatenation and no temporal skip connections on D-ResNet-18 architecture. For the sake of fairness, at each iteration all networks receive the current frame together with the two previous frames as input and apply a 3D convolution layer as the first operation. For summation and no temporal skip connection, 2D convolution layer is applied afterward, whereas for concatenation temporal skip connection, a 3D convolution layer is used since volumes are concatenated along the depth dimension. Although using a 3D convolution layer increases the number of parameters and floating point operations, concatenation achieves the best performance with a margin of $\sim$2.7\%.

We would like to note that summation does not bring any performance gain and even performs slightly worse than no temporal skip connection. We infer that this is due to the loss of temporal information after the summation operation.

It is also interesting to see that D-ResNet-18 with no skip connection achieves even better than conventional 3D~ResNet-18 architecture in Table \ref{tab:comparison_offline}. This contradicts the findings of \cite{tran2018closer}, where f-R2D achieves 1.3\% worse accuracy than R3D. Our only difference from f-R2D in \cite{tran2018closer} is that we apply a 3D convolution layer at the first convolution operation, which was enough to capture necessary motion information to outperform R3D. Besides, we can conclude that preserving temporal-resolution in the network (i.e. not applying temporal downsampling) increases classification performance, although this also increases the computation and memory load at inference time.

\begin{table*}[t!]
    \centering
    \begin{tabular}{lcccccc}
        \specialrule{.15em}{.1em}{.1em}
        \textbf{Model} & \textbf{Params} & \textbf{Cache Size} & \textbf{MFLOPs} & \textbf{Speed (ms)} & \textbf{St-Modeling} & \textbf{Accuracy (\%)}   \\ 
        \specialrule{.15em}{.1em}{.1em}
        3D-ResNet-18 \cite{kopuklu2019resource}        & 33.24M  & --   & 8323    & 3.00  & fc  & 57.65    \\
        3D-ResNet-50 \cite{kopuklu2019resource}        & 44.24M  & --   & 9835    & 5.46  & fc  & 63.00    \\
        3D-ResNet-101 \cite{kopuklu2019resource}       & 83.29M  & --   & 13664   & 7.04  & fc  & 64.18    \\
        \specialrule{.15em}{.1em}{.1em}
        D-ResNet-18      & 20.66M  & 0.97MB   & 747     & 0.33  & fc  & \phantom{aaaa} 61.41 \scriptsize \color{green}+3.76    \\
        D-ResNet-50      & 40.81M  & 1.95MB   & 1654    & 0.75  & fc  & \phantom{aaaa} 67.35 \scriptsize \color{green}+4.35    \\
        D-ResNet-101     & 69.83M  & 2.81MB   & 3077    & 1.46  & fc  & \phantom{aaaa} 68.78 \scriptsize \color{green}+4.60    \\
        \specialrule{.15em}{.1em}{.1em}
        D-ResNet-18      & 31.05M  & 0.94MB   & 845     & 0.33  & LSTM  & \phantom{aaaa} 62.02 \scriptsize \color{green}+4.37    \\
        D-ResNet-50      & 51.20M  & 1.92MB   & 1752    & 0.75  & LSTM  & \phantom{aaaa} 68.22 \scriptsize \color{green}+5.22   \\
        D-ResNet-101     & 80.22M  & 2.78MB   & 3175    & 1.46  & LSTM  & \phantom{aaaa} \textbf{69.17} \scriptsize \color{green}+4.99 \\
        \specialrule{.15em}{.1em}{.1em}
    \end{tabular}
    \caption{Comparison of D-ResNet architecture with conventional ResNet architecture over offline classification accuracy, number of parameters, computation complexity (FLOPs) at online operation on  the Kinetics-600 validation set. The cache size is calculated according to 32 bit floating point values for intermediate volumes and reported in megabytes (MB). For each architecture, the speed refers to single inference time measured using NVIDIA Titan XP GPU for a batch size of 8.}
	\label{tab:comparison_offline}
\end{table*}

\begin{figure*}
    \centering
    \subfigure[]{
    \begin{tikzpicture}
          \begin{axis}[
            width=8cm,
            height=4cm,
            xlabel=Segments,
            ylabel=Accuracy ($\%$),
            grid=both,
            legend columns=2,
            legend style={at={(0.5,0.3)},anchor=north}]
            \addplot coordinates {
                (1,  40.99)
                (2,  45.54)
                (3,  49.09)
                (4,  52.07)
                (5,  54.86)
                (6,  57.76)
                (7,  59.90)
                (8,  61.01)
                (9,  61.68)
                (10, 62.00)
            };
            \addplot [red,mark=triangle] coordinates {
                (1,  44.75)
                (2,  46.84)
                (3,  47.82)
                (4,  49.32)
                (5,  51.04)
                (6,  53.23)
                (7,  52.84)
                (8,  50.89)
                (9,  49.14)
                (10, 47.49)
            };
          \addlegendentry{$LSTM$}
          \addlegendentry{$fc$}
          \end{axis}
        \end{tikzpicture}} \quad \subfigure[]{
        \begin{tikzpicture}
        \begin{axis}[
        width=8cm,
        height=4cm,
        xlabel=Random erasing percentage ($\%$),
        ylabel=Accuracy ($\%$),
        grid=both,
        legend columns=2,
        legend style={at={(0.5,0.3)},anchor=north}]
        \addplot coordinates {
            (60,  56.17)
            (65,  55.12)
            (70,  53.62)
            (75,  52.26)
            (80,  49.04)
            (85,  47.00)
            (90,  38.22)
            (95,  29.74)
        };
        \addplot [red,mark=triangle] coordinates {
            (60,  56.06)
            (65,  55.08)
            (70,  53.36)
            (75,  50.89)
            (80,  47.03)
            (85,  39.25)
            (90,  21.84)
            (95,  1.53)
        };
      \addlegendentry{$LSTM$}
      \addlegendentry{$fc$}
      \end{axis}
    \end{tikzpicture}}
    \caption{Causality analysis of deployed spatiotemporal modeling mechanisms. In (a), videos are separated into ten equal segments and network outputs at each segment are averaged for $fc$ and $LSTM$. In (b), the network outputs at the middle parts of the videos are replaced with the Gaussian noise.}
    \label{fig:causality_analysis}
\end{figure*}
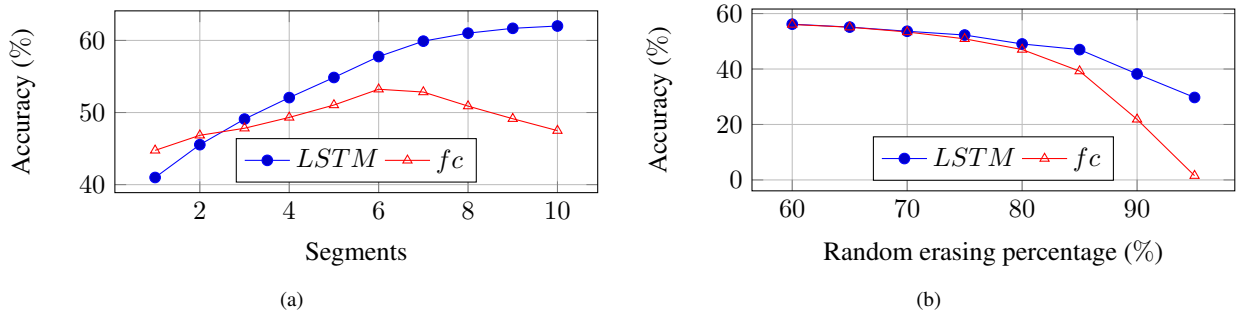

\vspace{0.25cm}
\noindent \textbf{\textit{Analysis of different spatiotemporal modeling mechanisms:}} We investigate the performance of applying fc, LSTM or GRU as spatiotemporal modeling mechanism. Table \ref{table:comp_st-modeling} shows the comparison of fc with LSTM and GRU for different number of hidden layers. Using multi-layer fc as spatiotemporal modeling mechanism slightly reduces performance. Both recurrent blocks perform better with 2 hidden layers while LSTM achieves the best performance. Hence, from this point onwards, we always use two layers for LSTM. However, we must note that RNNs in general need more parameters and FLOPs compared to fc. For example, single layer fc requires 4.9M parameters and 9.8M FLOPs, whereas 2 layer LSTM requires 15.3M parameters and 108.2M FLOPs for the settings explained in Section~\ref{sec:imp_details}.

\vspace{0.25cm}
\noindent \textbf{\textit{Effect of different clip lengths on training recurrent blocks:}} At the training of RNN blocks, clip length plays an important role on the final classification performance. We have investigated the effect of different clip lengths at training time on the classification performance for D-ResNet-18-lstm, as illustrated in Fig. \ref{fig:clip_lengths}. The results clearly show that a longer clip leads to higher classification accuracy. The reason is that LSTMs can learn the important/unimportant features and store/remove them in their cell state easier when they observe longer clips.

\begin{table}[t!]
    \centering
	\begin{adjustbox}{width=1.0\linewidth, center}
	\tabcolsep 3.80pt 
    \begin{tabular}{lccc}
        \specialrule{.15em}{.1em}{.1em}
        \textbf{Model} & \begin{tabular}[c]{@{}l@{}}\hspace{0.3cm}\textbf{Input} \\ \textbf{Resolution}\end{tabular} & \textbf{MFLOPS} & \textbf{Acc. (\%)}   \\ 
        \specialrule{.15em}{.1em}{.1em}
        3D-ShuffleNetV1 2.0x \cite{kopuklu2019resource}       & 112$\times$112   & 538    & 56.84  \\
        3D-ShuffleNetV2 2.0x \cite{kopuklu2019resource}       & 112$\times$112   & 438    & 55.17  \\
        3D-MobileNetV1 2.0x \cite{kopuklu2019resource}        & 112$\times$112   & 662    & 48.53  \\
        3D-MobileNetV2 1.0x \cite{kopuklu2019resource}        & 112$\times$112   & 561    & 50.65  \\
        3D-SqueezeNet \cite{kopuklu2019resource}              & 112$\times$112   & 926    & 40.52  \\
        3D-ResNet-18 \cite{kopuklu2019resource}        & 112$\times$112   & 8323  & 57.65    \\
        3D-ResNet-50 \cite{kopuklu2019resource}        & 112$\times$112   & 9835  & 63.00    \\
        3D-ResNet-101 \cite{kopuklu2019resource}       & 112$\times$112   & 13664  & 64.18    \\
        3D-ResNeXt-101 \cite{kopuklu2019resource}      & 112$\times$112   & 9652  & 68.30   \\
        I3D \cite{carreira2018short}                   & 224$\times$224   & 111331 & 71.90  \\
        Oct-I3D \cite{chen2019drop}  & 224$\times$224   & 25600  & 76.00 \\
        X3D-M \cite{feichtenhofer2020x3d}  & 224$\times$224   & 6200  & 78.80 \\
        X3D-XL \cite{feichtenhofer2020x3d}  & 224$\times$224   & 48400  & 81.90 \\
        SlowFast \cite{feichtenhofer2018slowfast}  & 224$\times$224   & 234000  & 81.80 \\
        \specialrule{.15em}{.1em}{.1em}
        D-ResNet-18      & 112$\times$112   & 845   & 62.02 \\
        D-ResNet-50      & 112$\times$112   & 1752  & 68.22 \\
        D-ResNet-101     & 112$\times$112   & 3175  & 69.17 \\
        \specialrule{.15em}{.1em}{.1em}
    \end{tabular}
    \end{adjustbox}
    \caption{Comparison of D-ResNet architecture with state-of-the-art on validation set of Kinetics-600 dataset. FLOPs are calculated for the inference to produce one output at online operation.}
	\label{tab:sota}
\end{table}

\vspace{0.25cm}
\noindent \textbf{\textit{Performance comparison of D-ResNet architectures with different depths:}} Comparative results are shown in Table \ref{tab:comparison_offline}. As usual, increasing network depth yields higher accuracies. Moreover, D-ResNet performs 74-90\%  less  computation  at  online  operation  while  achieving $\sim$5\% better classification accuracy compared to conventional ResNet models on Kinetics-600. This is since D3D uses the previous computations efficiently by caching the intermediate volumes of the network. The performance improvement is not also due to increased number of parameters since D-ResNet models have less parameters compared to conventional 3D-ResNet models. Only exception is D-ResNet-50 with LSTM, which has around 7M more parameters compared to \mbox{3D-ResNet-50.}  

\vspace{0.25cm}
\noindent \textbf{\textit{Required cache size for D-ResNet architectures:}} The amount of memory needed for cache to store intermediate volumes depends on (i) applied data precision (e.g., 32 bit, 16 bit or 8 bit floating point values), (ii) used input resolution and (iii) the number of stored intermediate volumes. Accordingly, we report the required cache size in Table~\ref{tab:comparison_offline}. As expected, deeper architectures store more intermediate volumes in the cache, hence requires more memory.

\vspace{0.25cm}
\noindent \textbf{\textit{Causality analysis of D-ResNet-18 architecture:}} The essential property of an online system is that the architecture should be causal. To validate the causality of the proposed D3D, we designed two tests. Firstly, we make a segment-level classification test, where we have divided input videos into ten equal parts and outputs are averaged within each segment. Fig. \ref{fig:causality_analysis}(a) shows the comparison of fc and LSTM. Since fc treats each clip independently and the middle parts of the videos are typically more informative, a bowed curve is achieved. On the other hand, LSTM stores the relevant features in its cell state over time, leading to increased accuracy with rising segment numbers. Keeping in mind that an entirely causal system should improve monotonically, D3D satisfies this criterion. Secondly, we have replaced the middle parts of the videos with the Gaussian noise and reported the performance comparison of fc and LSTM in Fig. \ref{fig:causality_analysis}(b). As we increase the erased percentage from the middle part of the videos, accuracy drops linearly for both fc and LSTM till 60\% erasure. If we keep increasing the erasure percentage, it becomes more and more important to use the information at the beginning and end of the videos jointly. Therefore, LSTM outperforms fc more and more as the erasure percentage increases. Specifically, with an erasure percentage of 95\%, fc achieves 1.53\% accuracy, whereas LSTM achieves 29.74\% accuracy.

\begin{table}[t!]
    \centering
    \begin{tabular}{lcc}
        \specialrule{.15em}{.1em}{.1em}
        \textbf{Model} & \textbf{St-Modeling} & \textbf{Accuracy (\%)}   \\ 
        \specialrule{.15em}{.1em}{.1em}
        3D-ResNet-18 \cite{kopuklu2019resource}        & fc  & 59.99    \\
        3D-ResNet-50 \cite{kopuklu2019resource}        & fc  & 68.14    \\
        3D-ResNet-101 \cite{kopuklu2019resource}       & fc  & 67.87    \\
        \specialrule{.15em}{.1em}{.1em}
        D-ResNet-18      & fc  & \phantom{aaaa} 63.01 \scriptsize \color{green}+3.02    \\
        D-ResNet-50      & fc  & \phantom{aaaa} 71.32 \scriptsize \color{green}+3.18    \\
        D-ResNet-101     & fc  & \phantom{aaaa} 70.97 \scriptsize \color{green}+3.10    \\
        \specialrule{.15em}{.1em}{.1em}
        D-ResNet-18      & LSTM  & \phantom{aaaa} 63.33 \scriptsize \color{green}+3.34    \\
        D-ResNet-50      & LSTM  & \phantom{aaaa} \textbf{71.50} \scriptsize \color{green}+3.36   \\
        D-ResNet-101     & LSTM  & \phantom{aaaa} 70.71 \scriptsize \color{green}+2.84 \\
        \specialrule{.15em}{.1em}{.1em}
    \end{tabular}
    \caption{Comparison of D-ResNet architecture with conventional 3D-ResNet architecture over video classification accuracy on the validation set of untrimmed ActivityNet dataset.}
	\label{tab:activitynet}
\end{table}

\vspace{0.25cm}
\noindent \textbf{\textit{State-of-the-art comparison:}} Although the motivation of this work is not beating the state-of-the-art, it is beneficial to see how well D3D architecture performs compared to other architectures. Accordingly, we have compared D3D architecture with the state-of-art architectures in Table~\ref{tab:sota}. 

We first note that FLOPs in Table~\ref{tab:sota} is calculated according to necessary number of floating point operations to produce one output at online operation. Therefore, FLOPs are calculated for the inference of `input clip' for all architectures other than \mbox{D-ResNet} family. Secondly, increased input resolution in general leads to better performance, but the computational complexity also increases quadratically. Thirdly, we would like to emphasize again that any 3D-CNN architecture in Table~\ref{tab:sota} can be converted to its dissected version. X3D-XL \cite{feichtenhofer2020x3d} and SlowFast \cite{feichtenhofer2018slowfast} achieves the best performance on Kinetics-600 dataset, but also require significantly high FLOPs at online operation. Therefore, their dissected version can satisfy both high accuracy and low computational complexity at online operation. If the resource efficiency is the main concern, the dissected version of resource efficient architectures such as 3D-ShuffleNetV1 2.0x \cite{kopuklu2019resource} would require very little computation with less than 100 MFLOPs.

\subsection{Video Activity Recognition On Untrimmed Datasets}

Experiment in Fig.~\ref{fig:causality_analysis}(b) shows that videos can be successfully recognized with D3D architectures with LSTM spatiotemporal modeling although some part of them contain activity-unrelated content. Therefore, we have investigated the performance of D3D on untrimmed ActivityNet dataset \cite{caba2015activitynet}. The videos in ActivityNet dataset are on average 117 seconds long and contain activities from 200 different classes. Therefore, with this dataset, we can also test the performance of D3D on longer videos. Similar to Kinetics-600, we have evaluated the performance of D3D on the validation set of ActivityNet dataset since test set is not publicly available. Used D-ResNet architectures are exactly same as Kinetics-600 experiments except for the last fc layer, which reduces the number of outputs according to the ActivityNet class number that is 200. All architectures are first pretrained on Kinetics-600 dataset and then fine-tuned on ActivityNet dataset.

\begin{table}[t!]
\centering
    \begin{tabular}{lcc}
        \specialrule{.15em}{.1em}{.1em}
        \textbf{Model} & \textbf{St-Modeling} & \textbf{Accuracy (\%)}   \\ 
        \specialrule{.15em}{.1em}{.1em}
        3D-ResNet-18 \cite{kopuklu2019resource}         & fc  & 93.34    \\
        D-ResNet-18       & fc  & \phantom{aaaaa} \textbf{94.58} \small\color{green}+1.24    \\
        \specialrule{.15em}{.1em}{.1em}
    \end{tabular}
    \caption{Comparison of dissected and conventional \mbox{ResNet-18} architectures on the Jester validation set. Both architectures take 16-frames input (downsampling of 2 is applied) with \mbox{112 $\times$ 112} spatial resolution.}
    \label{table:d3d_gesture}
\end{table}

In Table~\ref{tab:activitynet}, we compare the performance of D-ResNet with the conventional 3D-ResNet architectures. Firstly, ActivityNet dataset is not as large as Kinetics-600, hence deeper D-ResNet-101 performs worse than D-ResNet-50 due to overfitting. Secondly, D-ResNet achieves around 3\% better video classification accuracy compared to conventional 3D-ResNet architectures similar to Kinetics-600 dataset. We conjuncture that this is due to temporal resolution preserving property of D3D architectures. Lastly, using LSTM instead of fc does not improve accuracy as dramatic as Kinetics-600 dataset. This shows that increased video length is not that useful for ActivityNet dataset since it is an untrimmed dataset and the most of the video content does not contain auxiliary information for the correct classification of the video. Accordingly, we conclude that longer videos are favorable as long as all video content contributes to the correct prediction of the performed activity. The results of D-ResNet-18 with LSTM in Table~\ref{fig:causality_analysis}(a) justifies this argument.

\subsection{Gesture Recognition}

Gesture recognition can be viewed as a very similar task to action recognition task. In action recognition, although it is still necessary to capture motion patterns, the network especially needs to capture spatial patterns. For example, In the Kinetics-600 dataset, there are nine different ``eating something" classes where ``something" is one of ``burger, cake, carrot, chips, doughnut, hotdog, ice cream, spaghetti, watermelon". For the correct classification, the network must recognize the objects in the videos correctly. On the other hand, the spatial content in gesture videos are similar: A person in front of a camera performing a hand gesture. For the correct classification, the motion of the hand must be captured by the network.

To inspect the D3D architecture’s ability to capture motion patterns, we have experimented with the Jester dataset \cite{materzynska2019jester}, which is the largest available hand gesture dataset currently. Training details are kept exactly the same as previous settings. In Table \ref{table:d3d_gesture}, D-ResNet-18 achieves 1.24\% more classification accuracy than conventional 3D-ResNet-18.

\subsection{Video Person Re-Identification (ReID)}

Person Re-identification aims to match a queried data with its true owner in the gallery set. In video person ReID, both the query and gallery are person tracklets, which usually consist of a varying number of frames. Most state-of-the-art approaches leverage 2D CNN architectures for video ReID \cite{li2018diversity, li2019unsupervised, song2018mask}. However, 2D CNN architectures process individual frames independently, hence they cannot incorporate temporal information between frames. In this section, we demonstrate that our proposed D3D architecture can increase the performance over 2D CNNs.

\begin{table}[t!]
\centering
    \begin{tabular}{lcc}
        \specialrule{.15em}{.1em}{.1em}
        \textbf{Model} & \textbf{Accuracy (\%)} & \textbf{mAP}   \\ 
        \specialrule{.15em}{.1em}{.1em}
        2D-ResNet-50      & 80.8  & 69.0    \\
        D-ResNet-50       & \phantom{aaaa} \textbf{81.3} \small \color{green}+0.5 & \phantom{aaaa} \textbf{69.1}  \small \color{green}+0.1  \\
        \specialrule{.15em}{.1em}{.1em}
    \end{tabular}
    \caption{Comparison of our D-ResNet-50 architecture with 2D-ResNet-50 on the validation set of the MARS dataset.}
    \label{table:d3d_video_reid}
\end{table}

Our person ReID architecture is as follows. Given input video clips, a backbone network extracts features for each frame and these features are averaged to get final feature representing the given input clip. We utilized classification loss and triplet loss in order to train the network. For the classification loss, we consider person identities as category-level annotations and train a linear layer followed by a softmax operation to get class-conditional probabilities. Then, our classification loss $\mathcal{L_C}$ is the cross entropy error between the predicted classes and the ground truth classes. For the triplet loss $\mathcal{L_T}$, our data loader randomly selects $N$ video clips for each person, which is used for hard sample mining \cite{hermans2017defense}. The final loss is $\mathcal{L} = \mathcal{L_C} + \mathcal{L_T}$. The architecture is trained end-to-end using the final loss $\mathcal{L}$. In our experiments, we used $N=4$ and each clip contains 4 frames at training time. At test time, we have loaded all frames in person videos to get final video features.

In our experiments, we have used the MARS dataset \cite{zheng2016mars} for performance evaluation. For the backbone network in the architecture described above, we have compared the conventional 2D-ResNet-50 with our D-ResNet-50 architecture. Both models are inflated from ImageNet pretrained model. We trained the networks for 150 epochs using Adam optimizer with an initial learning rate 0.0003, which is divided by 10 every 60 epochs. In the MARS dataset, all person detections are already cropped, hence there is no pixel-wise correspondence at consecutive frames in tracklets. Therefore, we used single frames at the input of the D-ResNet-50 architecture. The comparative results are shown in Table~\ref{table:d3d_video_reid}. D-ResNet-50 architecture manages to capture discriminative motion information of identities, possibly gait-related information, which slightly increases the performance.


\begin{table}[t!]
\centering
\begin{adjustbox}{width=0.48\textwidth}
    \begin{tabular}{lccccc}
        \specialrule{.15em}{.1em}{.1em}
         \multirow{2}{*}{\textbf{Model}} & \multirow{2}{*}{\textbf{Fusion}}  & \multicolumn{4}{c}{\textbf{Accuracy (\%)}} \\  \cmidrule(lr){3-6}
          &    & \textbf{5}     & \textbf{25}    & \textbf{50}    & \textbf{100} \\
        \specialrule{.15em}{.1em}{.1em}
    2D-ResNet-18 & avg all & \textbf{93.18} & 93.66 & 93.64 & 93.66 \\
    D-ResNet-18 & avg all & 92.40 & \textbf{93.90} & 93.74 & 93.74 \\
    D-ResNet-18 & avg 5:end & 93.12 & 93.80 & \textbf{93.92} & \textbf{94.10} \\
        \specialrule{.15em}{.1em}{.1em}
    \end{tabular}%
    \end{adjustbox}
    \caption{Evaluation on YouTube Faces dataset resampled to different number of frames.}
    \label{table:d3d_video_faceid}
\end{table}

\subsection{Video Face Recognition}

In the domain of video face recognition, typical approaches \cite{liu2017sphereface, wang2018cosface,deng2019arcface} leverage the features obtained by training on big datasets containing still images followed by simple average pooling of the features without emphasis on the quality of every frame. More sophisticated approaches combine the feature extraction network to aggregate the features based on their importance with a feature aggregation network \cite{yang2017neural,xie2018multicolumn, xie2018comparator,zhong2018ghostvlad}. However, temporal information is discarded as frames are treated as an unordered set of faces. Compared to these approaches, our D3D architecture can cope with this task while only consisting of one single network.

For video face recognition task, we use VoxCeleb2 dataset \cite{chung2018voxceleb2} for pretraining the architectures and YouTube Faces dataset \cite{wolf2011YTF} to evaluate them. Before training the network, we preprocess the VoxCeleb2 dataset by extracting 3 frames per clip, which are aligned using facial landmarks extracted using the MTCNN \cite{zhang2016MTCNN} and cropped to $112 \times 112$ pixels. First, we pretrain a 2D-ResNet-18 with a 256-dimensional bottleneck layer on single image recognition on the VoxCeleb2 dataset using cross entropy loss with Adam optimizer, 50\% dropout, an initial learning rate of 0.05 and a batch size of 100 for 50 epochs. We decided against pretraining on a bigger dataset containing still images, as otherwise, the adaption to D-ResNet-18 gets overshadowed by the dataset change. For training the D-ResNet-18, we inflate the weights of the 2D-ResNet-18 and finetune using 5 frames per sample and a frame at the input with a lower learning rate of 0.01 and additional motion blur data augmentation for 1 epoch. Apart from these changes, parameters are identical to the pretraining. Our experiments showed that motion blur data augmentation did not improve the accuracy of the 2D-ResNet-18, whereas it improves accuracy while finetuning the D-ResNet-18.

We evaluated our approach on the YouTube Faces dataset following the standard verification protocol. We computed the Euclidean distance after l2-normalization and taking the average of the features. The preprocessing is done similarly to the VoxCeleb2 dataset. However, we resample the videos  to obtain a fixed number of frames to show the dependency of the accuracy on the number of frames as shown in Table \ref{table:d3d_video_faceid}. In contrast to the 2D-ResNet-18, our D-ResNet-18 continues to improve with increasing number of frames. We also evaluated discarding the first four features in the average due to cache initialization (denoted by avg 5:end), which resulted in another minor improvement. Note that for 5 frames \textit{avg 5:end} is equal to taking only the last feature, which is substantially higher than the accuracy of the 2D-ResNet-18 for a single frame per video ($88.78\%$). This demonstrate that our network is capable of propagating useful information through time.

\section{Conclusion}
\label{sec:con}

In this work, we have addressed the computational complexity drawback of 3D-CNNs and proposed a novel Dissected 3D-CNN (D3D) architecture. The D3D architecture caches the intermediate volumes of the network and propagates them for future calculation, which reduces the computation around 77-90\% during online operation for \mbox{D-ResNet} family. Besides reducing complexity during online operation, D-ResNet family achieves $\sim$5\% higher classification accuracy compared to classical ResNet family on Kinetics-600 dataset. We believe that this performance improvement arises since D3D networks are temporal resolution preserving and produce frame level fine-grained features. In this work, only ResNet family is converted to its dissected version and evaluated. However, any CNN architecture can be converted to its dissected version for efficient online video processing. The proposed D3D architecture successfully models temporal information and can be employed at any video based computer vision task. In our experiments, we have successfully validated the effectiveness of D3D architecture on five different vision tasks: activity/action recognition on trimmed and untrimmed datasets, gesture recognition, video person re-identification and video face recognition. For all these tasks, D3D consistently improves the performance. We believe that the D3D architecture will be actively used in many other video based tasks by the vision community.
\vspace{0.3cm}
\section*{Acknowledgements}
\label{sec:ack}

We gratefully acknowledge the support by the Deutsche Forschungsgemeinschaft (DFG) under Grant No. RI \mbox{658/25-2}. We also acknowledge the support of NVIDIA Corporation with the donation GPUs used in this study.

{\small
\bibliographystyle{ieee_fullname}
\bibliography{egbib}
}

\end{document}